# THE IMPACT OF DATA VOLUME ON PERFORMANCE OF DEPP LEARNING BASED BUILDING ROOFTOP EXTRACTION USING VERY HIGH SPATIAL RESOLUTION AERIAL IMAGES

*Hongjie He[1], Ke Yang[2*], Yuwei Cai[1], Zijian Jiang[1], Qiutong Yu[1], Kun Zhao[1], Junbo Wang[1], Sarah Narges Fatholahi[1], Yan Liu[1], Hasti Andon Petrosians[1], Bingxu Hu[1], Liyuan Qing[1], Zhehan Zhang[1], Hongzhang Xu[1], Siyu Li[1], Kyle Gao[1], Linlin Xu[2], Jonathan Li[1,2*]*

[1]Department of Geography and Environmental Management, University of Waterloo, Waterloo, ON N2L 3G1, Canada
[2]Department of Systems Design Engineering, University of Waterloo, Waterloo, ON N2L 3G1, Canada
*Corresponding authors: ke.yang@uwaterloo.ca (K. Yang) and junli@uwaterloo.ca (J. Li)

## ABSTRACT

Building rooftop data are of importance in several urban applications and in natural disaster management. In contrast to traditional surveying and mapping, by using high spatial resolution aerial images, deep learning-based building rooftops extraction methods are efficient and accurate. Although more training data is preferred in deep learning-based tasks, the effect of data volume on building extraction models is underexplored. Therefore, the paper explores the impact of data volume on the performance of building rooftop extraction from very-high-spatial-resolution (VHSR) images using deep learning-based methods. To do so, we manually labelled 0.12m spatial resolution aerial images and perform a comparative analysis of models trained on datasets of different sizes using popular deep learning architectures for segmentation tasks, including Fully Convolutional Networks (FCN)-8s, U-Net and DeepLabv3+. The experiments showed that with more training data, algorithms converged faster and achieved higher accuracy, while better algorithms were able to better mitigate the lack of training data.

## 1. INTRODUCTION

Building maps are the essential data for urban planning and management, population estimation, urban cadastral management, insurance, and natural disaster management [1]. Given their importance, high definition (HD) maps generation, especially those with accurate building locations and shape information, has drawn much attention [2]. Traditional methods, such as surveying and mapping can accurately generate the building rooftops maps; these are however resource intensive. Furthermore, time cost is limiting the feasibility of traditional methods.

With the development of imaging techniques and sensors, high spatial resolution (HSR) and very high spatial resolution (VHSR) images have become widely used in HD mapping. Building rooftop extraction from HSR/VHSR images has proven to be efficient compared to traditional methods. Different types of information, such as morphological information, texture information and shape information, are employed to extract building rooftops accurately [3]. Although LiDAR data can be used to extract building information accurately with self-provided height information, they require more pre-processing. Therefore, compared to images, LiDAR data may not be the best choices for building rooftop extraction. In addition, with stereo images, height information is also available [4].

Recently, once prevailed object-based image (OBIA) or geographic object-based image analysis (GEOBIA) methods have been replaced by pixel based or patch-based methods with deep learning techniques [5]. For OBIA, segmentation must be done before classification to construct super-pixels or objects. However, parameters selection in the segmentation stage is time intensive. Furthermore, OBIA can be combined with traditional machine learning methods but not deep learning methods, which further limits its application since the latter have overtaken classical methods in performance [6]. Therefore, pixel-based methods or patch-based methods combined with deep learning techniques are the current state-of-the-art in building rooftop extraction because of their high performance.

New deep learning methods [7] achieved high performance in remote sensing. However, deep learning-based image segmentation methods are known to be data intensive. Although it is known that with large scale training data, models can be more generalizable and achieve higher accuracy, we are not aware of much prior work which explore the impact of data volume on deep learning-based image segmentation tasks in remote sensing applications. Therefore, in this paper, we analyze the impact changing the quantity of training data in building rooftops extraction, one of the image segmentation tasks in remote sensing.

In next section, we introduce the dataset and methods used in this paper. The segmentation results and evaluation are presented in section 3. We conclude the work with our findings in section 4. The contribution of the work includes:

1. We describe the process of ab-initio deep learning-based building rooftops extraction starting from the creation of a custom dataset.
2. We provide the comparison of the performance of models with different training data size to show the influence of data size on different architectures.

## 2. DATASET AND METHODS

### 2.1 Dataset

From images covering Kitchener-Waterloo region, because of time limitation, we manually edit 36 images of size 8350×8350 with a spatial resolution of 0.12m. The original images and the location of the study area are shown in Fig.1. Considering the memory footprint and computation requirement for the forward and backward propagation of large images, we further crop and pad the dataset into 10404 small tiles of size 512×512. (Same dataset is also used in paper entitled with "A Comparative Study of Deep Learning Approaches to Rooftop Detection in Aerial Images".) A sample Full Image and sample Tile are provided in Fig.2.

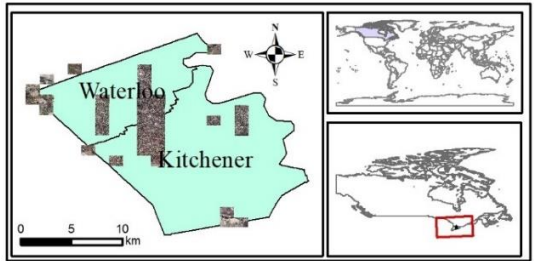

Fig.1. Kitchener-Waterloo area, Canada

A total of 36 8350×8350 images are cropped without overlap and padded into 10404 small tiles of size 512×512. Those tiles are further divided into 6069 and 4335 pairs of images for training and testing, respectively.

### 2.2 Methods

To analyze the impact of data volume on the performance of building rooftops extraction, we compare models trained on the full dataset, as well as reduced datasets with 75% and 50% of original images and labels. Eventually, 4550 pairs of images were generated for 75% size training set. A total of 3035 pairs of images are generated for 50% size training set.

For a fair analysis, we selected three widely used deep learning-based segmentation architectures in remote sensing. They are FCN-8s, U-Net and DeepLabv3+, which can extract rooftop masks from images in an end-to-end manner pixel-wisely. For the detailed description of these architecture, readers are referred to [8,9,10]. For each architecture and training set size, we train a separate model using fixed hyperparameters. To evaluate the performance of the models mentioned above, commonly used metrics, including average accuracy or, IoU (Intersection of Union), mIoU (mean IoU), precision, recall and F1 scores, are employed for performance assessment. Readers are referred to [11] for the definition of average accuracy, IoU, precision, recall and F1 score. mIoU is defined in this context as the mean of positive objects' (buildings) IoU and negative objects' (background) IoU. Inference time is only affected by model architecture, image size and hardware, we omit its evaluation in this work. The flowchart of the methodology is provided in Fig.3.

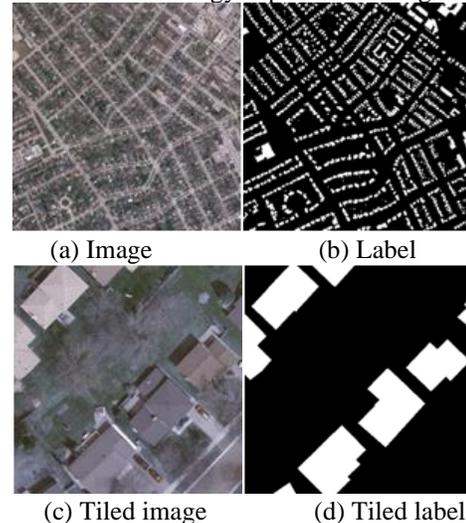

(a) Image          (b) Label

(c) Tiled image      (d) Tiled label

Fig. 2. Sample of digitalized images. (a) and (b) have the size of 8350×8350. (c) and (d) have the size of 512×512.

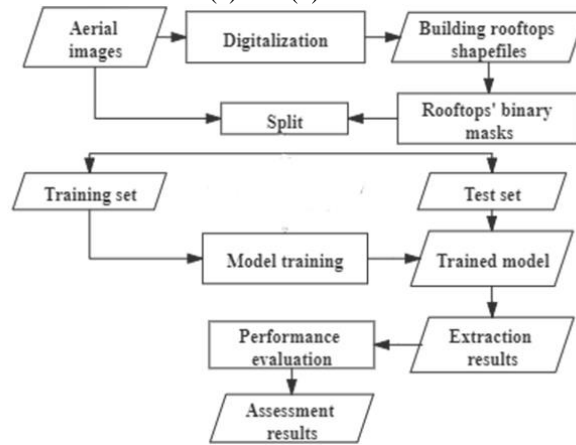

Fig. 3. Flowchart of the methodology

## 3. RESULTS AND EVALUATION

For consistency, all models were trained using the Adam optimizer [12] with a learning rate of 1E-4. The models were trained for 100 epochs with 5 tiles per batch on a GeForce RTX 2080ti GPU with CUDA 10.2.

### 3.1. Qualitative Results

Table 1 show the segmentation results of our models and provide a visual comparison of the effect of different architectures and data quantity. As can be seen, the

Table 1. Examples of segmentation results

| Architectures | Image | Ground Truth | 100% Data | 75% Data | 50% Data |
|---|---|---|---|---|---|
| DeepLabv3+ | | | | | |
| U-Net | | | | | |
| FCN-8s | | | | | |

Note: 100% data, 75% data and 50% data here indicate the proportion of training dataset used.

performance difference between different architecture is more significant than the difference between training set size. DeepLabv3+ achieves near perfect rooftop segmentation when trained on the full dataset. DeepLabv3+ trained on half of the dataset still achieves better segmentation results than FCN-8s trained on the full dataset. To properly quantify this effect, we evaluate the impact of changing training set size on different architectures using performance metrics.

## 3.2. Quantitative Results

The impact of training data size on the performance of the chosen architectures is analyzed in this section. The experiment shows that when training data size increases, models converged faster. Since all three models show same trend, we analyzed DeepLabv3+ in detail as an example (Fig. 4). One interesting effect we observed is the increased stability of the convergence with larger training sets, especially when approaching the end of the training stage. The speed of convergence, as well as the stable convergence at the end of training, may be attributed to more backpropagation steps in each epoch for larger training sets.

By evaluating various models using performance metrics, the impact of data volume is further explored. As can be seen from Table 2, with more training data, algorithms achieve higher performance. It is worth noting that DeepLabv3+'s performance is affected less by data volume than U-Net's and FCN-8s'. In particular, when removing 25% of the training data, DeepLabv3+ only suffers an average accuracy loss of 0.4%, whereas FCN-8s and U-Net suffer 1.8% and 2.8% average accuracy loss respectively.

## 4. CONCLUSION

In this paper, the impact of data volume to the performance of deep learning building rooftops extraction models is explored. Three widely used deep learning architectures trained on manually labelled aerial image datasets of different sizes are evaluated and analyzed. From experimental results, we can conclude that with more training data, better extraction results can be obtained. This effect is less apparent in advanced architectures, where it is not necessarily noticeable by visual inspection. This is supported by the performance metrics which show less performance loss when reducing the training set size for DeepLabv3+ when compared to FCN-8s. Future research directions include the analysis of different models, as well as the investigation of overfitting datasets of various sizes.

Table 2. Performance evaluation with different training data size (%)

| Models | Average accuracy | IoU | mIoU | Precision | Recall | F1-score |
|---|---|---|---|---|---|---|
| **FCN-8s - 100% *data*** | 80.8 | 37.7 | 58.0 | 41.3 | 80.9 | 54.7 |
| **FCN-8s - 75% *data*** | 79.0 | 34.4 | 55.4 | 38.4 | 76.7 | 51.2 |
| **FCN-8s - 50% *data*** | 69.9 | 28.1 | 47.0 | 30.0 | 82.2 | 43.9 |
| **U-Net - 100% *data*** | 93.3 | 64.6 | 78.4 | 72.4 | 85.6 | 78.5 |
| **U-Net - 75% *data*** | 90.5 | 56.4 | 72.8 | 62.7 | 84.8 | 72.1 |
| **U-Net - 50% *data*** | 81.4 | 40.5 | 59.6 | 42.8 | 88.1 | 57.7 |
| **DeepLabv3+ - 100% *data*** | 93.7 | 65.8 | 79.3 | 75.3 | 83.9 | 79.4 |
| **DeepLabv3+ - 75% *data*** | 93.3 | 63.6 | 78.0 | 74.5 | 81.4 | 77.8 |
| **DeepLabv3+ - 50% *data*** | 91.9 | 52.4 | 71.7 | 76.6 | 62.3 | 68.7 |

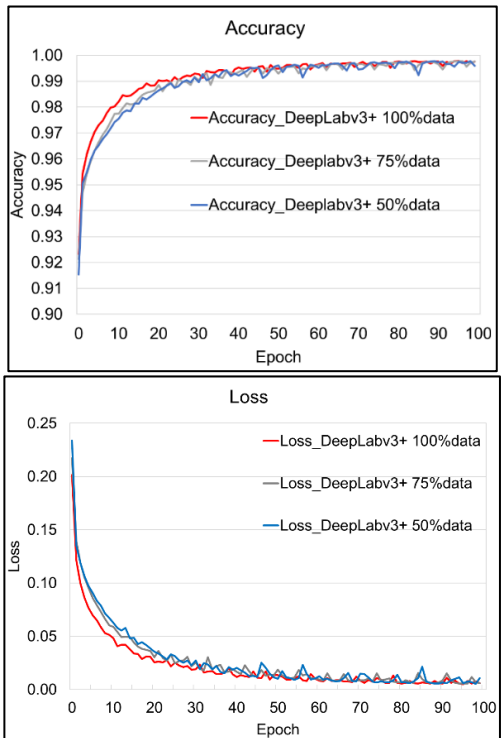

Fig.4. Accuracy and Loss training curves for DeepLabv3+

## 5. AKNOWLEDGEMENTS

The first author would acknowledge the China Scholarship Council for their support via a doctoral scholarship (No. 201906180088). We also acknowledge the Geospatial Centre at University of Waterloo for providing aerial images.